\documentclass[twoside,11pt]{article}

\usepackage[preprint,hyperref,abbrvbib]{jmlr2e}

\usepackage{amsmath}
\usepackage{bm}
\usepackage[dvipsnames]{xcolor}
\usepackage{enumitem}
\usepackage{xfrac}

\usepackage{tikz}
\usetikzlibrary{shapes,arrows,calc,positioning}
\usepackage{caption}
\usepackage{subcaption}
\usepackage[section]{placeins}

\DeclareMathOperator*{\argmax}{arg\,max}

\DeclareMathOperator*{\Var}{Var}
\DeclareMathOperator*{\sgn}{sgn}

\usepackage{lastpage}
\jmlrheading{25}{2024}{1-\pageref{LastPage}}{4/24}{}{21-0000}{Bart M.N. Smets, Peter D. Donker and Jim W. Portegies}

\ShortHeadings{Semiring Activation in Neural Networks}{Smets, Donker and Portegies}
\firstpageno{1}
\usepackage{color}

\begin{document}

\title{Semiring Activation in Neural Networks}

\author{\name Bart M.N. Smets \email b.m.n.smets@tue.nl \\
    \name Peter D. Donker \email p.d.donker@student.tue.nl \\
    \name Jim W. Portegies \email j.w.portegies@tue.nl \\
    \addr Department of Mathematics and Computer Science\\
    Eindhoven University of Technology, the Netherlands
}

\editor{My editor}

\maketitle

\begin{abstract}
We introduce a class of trainable nonlinear operators based on semirings that are suitable for use in neural networks.
These operators generalize the traditional alternation of linear operators with activation functions in neural networks.
Semirings are algebraic structures that describe a generalised notation of linearity, greatly expanding the range of trainable operators that can be included in neural networks.
In fact, max- or min-pooling operations are convolutions in the tropical semiring with a fixed kernel.

We perform experiments where we replace the activation functions for trainable semiring-based operators to show that these are viable operations to include in fully connected as well as convolutional neural networks (ConvNeXt). 
We discuss some of the challenges of replacing traditional activation functions with trainable semiring activations and the trade-offs of doing so.
\end{abstract}


\section{Introduction}

Neural networks come in a large variety of types and architectures, one common characteristic the majority of them share is the alternation between trainable linear operations (or affine operations if one includes bias) on the one hand and non-trainable nonlinear operators in the form of a scalar activation function on the other hand. 
Sometimes a multi-variate nonlinear function like max/min-pooling is also used, but this is again a fixed, non-trainable function similar in nature to a scalar activation function.
Even in the case of transformers \citep{vaswani2017attention} one trains linear maps that are then composed in a fixed manner through the inner product and a soft-max function.
While this binary setup is the de facto standard in machine learning, exceptions exist.
These exceptions can be roughly classified into three classes: trainable activation functions, non-standard neurons and morphological.
We will briefly discuss these three approaches. 
For a comprehensive overview of trainable activation functions and non-standard neurons see \citet{apicella2021survey}.

\paragraph{Trainable activation functions.}
The most straightforward way of obtaining a trainable nonlinearity is using an activation function that has one or more parameters affecting it and train these parameters as part of the neural network.
This is typically done either by adding some shape parameter to an existing fixed activation function or building the activation function as an ensemble of some fixed basis functions and have the ensemble coefficients be trainable parameters.
An example of the first kind is the \emph{swish} function \citep{ramachandran2017searching} defined as
\begin{equation*}
    \operatorname{swish}_\alpha (x) 
    := 
    x \cdot \operatorname{sigmoid}(\alpha \cdot x)
    ,
\end{equation*}
where $\alpha \in \mathbb{R}$ is a parameter that can either be trained or constant.
Note that for $\alpha=1$ the swish function reduces to the SiLU \citep{elfwing2018sigmoid} and for $\alpha\to\infty$ reduces to the familiar ReLU.

An example of the second kind is the \emph{adaptive blending unit} \citep{sutfeld2018adaptive} given by
\begin{equation*}
    \operatorname{ABU}(x)
    :=
    \sum_{i=1}^k \alpha_i \cdot f_i(x)
    ,
\end{equation*}
where the $\alpha_i$'s are the trainable parameters and the functions $f_i$ are a selection of fixed activation functions such as tanh, ReLU, id, swish, etc.
In the original study the $\alpha_i$ parameters were initialized with $\frac{1}{k}$ and constrained using a normalization scheme.

As \citet{apicella2021survey} notes, in most cases these trainable activation functions can also be expressed (or approximated) as a small feed-forward neural network with fixed activation functions.
This is not surprising since the fixed activation functions remain the core building blocks.
As a consequence, the same benefit of using these trainable activation functions can be achieved by simply making the neural network deeper.

\paragraph{Non-standard neurons.}
The alternative to making the activation function trainable is doing away with the standard neuronal model (linear map followed by an activation function) altogether.
A basic example is a \emph{maxout network} \citep{goodfellow2013maxout} where each neuron has a number of linear maps with scalar codomain and the final output is the maximum of the group.

A more elaborate non-standard neuron is used in an \emph{morphological neural network} as introduced by \citet{ritter1996introduction}. 
Here the idea is to replace the inner linear combination in the classic neuron given by
\begin{equation*}
    y_j = 
    \sigma \left(
        \textstyle\sum_{i=1}^n w_{ij} \ x_i + b_j
    \right),
\end{equation*}
where $\sigma : \mathbb{R} \to \mathbb{R}$ is an activation function , with a \emph{tropical} combination given by:
\begin{equation*}
    y_j = 
    \sigma \left(
        \max \left\{ w_{ij} +  x_i  \ \middle|\ i = 1 \ldots n \right\}
        +
        b_j
    \right)
    .
\end{equation*}
and where $\sigma$ is again a choice of scalar activation function.
More examples of non-standard neurons can be found in \citet{apicella2021survey}.

\paragraph{Morphological.}
Mathematical morphology is a theory and set of techniques for analyzing and processing geometric features, most commonly in images \citep{serra1982image}.
The general technique is similar to convolution where one move a kernel along an input image and multiply and integrate to generate an output.
Instead one calls the kernel the \emph{structuring element} and instead of multiplying one add or subtracts and then takes the maximum or minimum instead of integrating.
This operation is also referred to as \emph{morphological convolution}, and can be thought of as a nonlinear version of the familiar linear convolution.
The use of morphological techniques in deep learning can be considered a special case of non-standard neurons, but it is specific enough to warrant separate discussion.

The earliest work we are aware of is the \emph{PConv} operator in \citet{masci2013learning}, here the morphological convolution is not implemented directly but approximated through counter-harmonic means, which is also the approach taken in \citet{mellouli2017morph}.
Later work uses other approximations like soft maximum and minimum in \citet{shih2019development}.
Both approaches have some trouble when executed in floating point arithmetic, that can however be ameliorated to some degree by a smart choice of bias as in \citet{shen2019deep}.
Direct computation of morphological convolution is also possible as is used in \emph{PDE-based CNNs} by \citet{smets2023pde}.

Our work looks at linear convolution and morphological convolution as two special cases of a more general family of operators where a particular choice of codomain algebra and domain translation equivariance naturally yields a convolution type operator.

\paragraph{Our approach.}

Instead of looking at deep neural networks as consisting of layers of neurons, we can also take the view that we are alternating between linear and nonlinear operators.
Sure, we usually choose the nonlinear operator to be a scalar activation function and 
let the linear operators be the trainable part but that is but one possible design choice.

Suppose we could place the nonlinear operator on an equal footing to the linear one, what would this look like?
Clearly, the nonlinear operator needs some sort of structure similar to the linear operators to make this work.
If \(A:V \to W\) is some map between vector spaces \(V\) and \(W\) then we say it is linear if
\[
    A(a v_1 + b v_2) = a A(v_1) + b A(v_2)
\]
for all  \(v_1,v_2 \in V\) and scalars \(a\) and \(b\).
We could require a similar structure of a nonlinear operator \(B:X \to Y\), so that for some binary operations \(\oplus\) and \(\odot\) we have
\begin{equation}
    B(a \odot x \oplus b \odot y) 
    = 
    a \odot B(x) \oplus b \odot B(y)
    \label{eq:quasilinear}
\end{equation}
for all \(x_1,x_2 \in X\) and scalars \(a\) and \(b\).
The spaces \(X\) and \(Y\) would have to be spaces where these operations make sense of course.
If this allows for the nonlinear operator \(B\) to be written as a matrix, similar to \(A\) then we could train \(B\) in the same way we train \(A\) and effectively have linear and nonlinear operators on the same footing.

This then will be our approach, look at a class of nonlinear operators that are \emph{quasilinear} as in \eqref{eq:quasilinear} and then instead of building neural networks with trainable linear (or affine) operators \(A_i\) and activation functions \(\sigma\) as
\begin{equation*}
    A_L \circ \sigma \circ A_{L-1} \circ \cdots \circ A_2 \circ \sigma \circ A_1,
\end{equation*}
we build it with linear operators \(A_i\) and nonlinear operators \(B_i\) as 
\begin{equation*}
    A_L \circ B_{L-1} \circ A_{L-1} \circ \cdots \circ A_2 \circ B_1 \circ A_1
    ,
\end{equation*}
where both types are trainable.
We will refer to this idea as \emph{semiring activation}.

An additional property of this approach is that it is entirely compatible with the notion of \emph{geometric equivariance}: the property of a model where when a certain geometric transformation is applied to the input (say a translation is applied to an image input) then the output of the model should undergo a corresponding transformation (if the output is also an image, that image should also be translated, if the output is a classification label then the label should not change, i.e. be invariant).

The original impetus of the approach we take in this work came from our previous research into \emph{PDE-based equivariant CNNs} in \citet{smets2023pde}.
In that work we also built neural networks without activation functions and used trainable nonlinear operators based on certain PDE solvers.
The PDE solvers we considered -- while nonlinear -- did have a \emph{quasilinear} property of the same kind as we will discuss in this work. 
The question we asked ourselves was how much of the performance of the PDE-G-CNNs in \citet{smets2023pde} is due to the PDE structure? Or, is much of the performance explained by just the semiring structure of the nonlinearities?
In this work we examine how far we can get using just semiring based nonlinearities.

\section{Quasilinear operators from semirings}

To construct quasilinear operator as in \eqref{eq:quasilinear} we need our generalized addition \(\oplus\) and generalized multiplication \(\odot\) to share some behaviour with the conventional ``\(+\)" and ``\(\cdot\)". 
Specifically, we will require them to form a semiring.

\begin{definition}[Semiring]
    A semiring is a set \(R\) equipped with two binary operations \(\oplus\) and \(\,\odot\,\), called addition and multiplication, so that
    \begin{enumerate}[label=(\roman*), topsep=2pt, noitemsep]
    \item addition and multiplication are associative,
    \item addition is commutative,
    \item addition has an identity element \(0_R\),
    \item multiplication has an identity element \(1_R\),
    \item multiplication distributes over addition:
            \begin{equation*}
                a \odot (b \oplus c) = a \odot b \oplus a \odot c
                \quad\text{and}\quad
                (a \oplus b) \odot c = a \odot c \oplus b \odot c,
            \end{equation*}
    \item multiplication with \(0_R\) annihilates: \( 0_R \odot a = a \odot 0_R = 0_R\).
    \end{enumerate}
\end{definition}
Additionally, if \(a \oplus a = a\) we say the semiring is \emph{idempotent} and if \(a \odot b=b\odot a\) we say the semiring is \emph{commutative}.
As is conventional we let multiplication take precedence over addition: \(a \oplus b \odot c = a \oplus (b \odot c)\).

In the same way we construct linear maps through matrix multiplication we can now construct a quasilinear map.
Let \(R\) be some commutative semiring, \(x \in R^n\) and \(B \in R^{m \times n}\) and define
\begin{equation}
    B \odot x =
    \begin{bmatrix}
        b_{11} & \cdots & b_{1n}
        \\
        \vdots & \ddots & \vdots
        \\
        b_{m1} & \cdots & b_{mn}
    \end{bmatrix}
    \odot
    \begin{bmatrix}
        x_1 \\ \vdots \\ x_n
    \end{bmatrix}
    :=
    \begin{bmatrix}
        b_{11} \odot x_1 \oplus \cdots \oplus b_{1n} \odot x_n
        \\
        \vdots
        \\
        b_{m1} \odot x_1 \oplus \cdots \oplus b_{mn} \odot x_n
    \end{bmatrix}
    .
    \label{eq:generalized-matrix-vector}
\end{equation}
Which is nothing more than the usual matrix-vector multiplication operation when we take the linear semiring \(R=(\mathbb{R},+,\,\cdot\,)\).
When we further overload the symbols \(\oplus\) and \(\odot\) to component-wise addition resp. scalar multiplication in \(R^n\), i.e.:
\begin{equation*}
    a \odot \begin{bmatrix} x_1 \\ \vdots \\ x_n \end{bmatrix}
    :=
    \begin{bmatrix} a \odot x_1 \\ \vdots \\ a \odot x_n \end{bmatrix}
    \quad
    \text{and}
    \quad
    \begin{bmatrix} x_1 \\ \vdots \\ x_n \end{bmatrix}
    \oplus
    \begin{bmatrix} y_1 \\ \vdots \\ y_n \end{bmatrix}
    :=
    \begin{bmatrix} x_1 \oplus y_1 \\ \vdots \\ x_n \oplus y_n \end{bmatrix},
\end{equation*}
then it can be verified that \(B\) satisfies
\begin{equation}
    B \odot (a \odot x \oplus b \odot y)
    =
    a \odot (B \odot x) \oplus b \odot (B \odot y)
    .
    \label{eq:quasilinear2}
\end{equation}
Which makes \(x \mapsto B \odot x\) satisfy \eqref{eq:quasilinear}, we say it is a \emph{quasilinear operator with respect to} \(R\).
Adding a bias can also be done in this setting: \(x \mapsto B \odot x \oplus c\) with \(c \in R^m\).

\begin{remark}[Semimodules and their homomorphisms]
    In algebraic terms if we have a semiring \(R\) then \(R^n\) equipped with the component-wise addition \(\oplus:R^n \times R^n \to R^n\) and scalar multiplication \(R \times R^n \to R^n\) from above forms a left \(R\)-\emph{semimodule}.
    A semimodule is a generalization of the concept of a vector space, where the underlying set of scalars is a semiring but not necessarily a field like \(\mathbb{R}\).
    All fields are semirings and all vector space are semimodules but not the other way around.
    Quasilinearity as in \eqref{eq:quasilinear} and \eqref{eq:quasilinear2} can then be understood as a homomorphism between semimodules.
\end{remark}

\section{Logarithmic and tropical semirings}

Semirings, being fairly general, come in a large variety.
For our current purpose we are only interested in semirings that have the real numbers, possibly extended with \(\pm\infty\), as their set.
This way the new quasilinear operators can coexist with the linear operators since each can deal with the output of the other.
The new addition and multiplication operations should also be easy enough to compute in practice so as to not incur an unreasonable performance penalty.

\paragraph{Logarithmic semirings.} The first semirings we consider is the family of \emph{logarithmic semirings}.
Let \(\mu \neq 0\) and define
\begin{equation*}
    a \oplus_\mu b := \frac{1}{\mu} \log\left( e^{\mu a} + e^{\mu b} \right)
    \quad \text{and} \quad
    a \odot b := a + b.
\end{equation*}
Here we adopt the convention that \(e^{\infty}=\infty\), \(e^{-\infty}=0\) and correspondingly \(\log{(\infty)}=\infty\) and \(\log{(0)}=-\infty\).
Then for \(\mu>0\) we have that \(R_{\log}^\mu:=(\mathbb{R} \cup \{-\infty\},\oplus_\mu,+)\) forms a commutative semiring.
Associativity, commutativity and distributivity are easy to check.
The identity element for addition is \(0_R = -\infty\) and for multiplication \(1_R=0\).
The annihilation axiom is also satisfied since \(0_R \odot a = -\infty + a = -\infty = 0_R\).

Similarly for \(\mu < 0\) we have that \(R_{\log}^\mu:=(\mathbb{R} \cup \{\infty\},\oplus_\mu,+)\) is a commutative semiring except that now the identity element for the addition is \(0_R=\infty\).

\paragraph{Tropical semirings.} The second family we consider is that of the \emph{tropical semirings} \(R_{\max}:=(\mathbb{R} \cup \{-\infty\},\max,+)\) and \(R_{\min}:=(\mathbb{R} \cup \{\infty\},\min,+)\), also called the \emph{max-plus semiring} and \emph{min-plus semiring} respectively.
For the max-plus semiring, \(-\infty\) is the additive identity since \(\max(a,-\infty)=a\) while for the min-plus semiring \(\infty\) is the additive identity, \(0\) is the multiplicative identity for both.
These two semirings are isomorphic under negation \(x \mapsto -x\).
Both the tropical semirings are commutative and idempotent since \(\max(a,a)=\min(a,a)=a\).

\begin{remark}
Morphological neural networks as in \citet{ritter1996introduction} replace the linear combination of the neuron with what we could now call a tropical combination, indeed \(\max_{i=1\ldots n} \{w_{i}+x_i\}\) is a quasilinear function in the max-plus semiring and can be written as
\begin{equation*}
    y_j = \bigoplus_{i=1}^n w_{ij} \odot x_i
    ,
\end{equation*}
with respect to the tropical \(\oplus\) and \(\odot\).
Our approach is distinct in that 1) we do not aim to replace the linear combination but rather the activation function while keeping the linear combination, 2) have a variety of semirings to consider.
\end{remark}

\section{In fully connected networks}

To check the viability of our proposed scheme we first run a set of experiments with small fully connected networks on a variety of datasets.
The datasets for this series of experiments are:
\begin{itemize}
    \item the classic iris dataset from \cite{anderson1936species}, available at \\
    \url{https://www.kaggle.com/datasets/uciml/iris]},
    \item the heart disease dataset available at \\
    \url{https://www.kaggle.com/datasets/johnsmith88/heart-disease-dataset},\footnote{The provenance of this dataset is complicated and its contents are of questionable value for predicting heart disease, see \cite{simmons2021investigating} for an investigation into this dataset. For our purpose of comparing network architectures this dataset is still perfectly serviceable to see how well models deal with poor data.}
    \item the circles and spheres dataset from \cite{naitzat2020topology}, available at \\
    \url{https://github.com/topnn/topnn_framework},
    \item the FashionMNIST dataset from \cite{xiao2017online}, available at \\\url{https://github.com/zalandoresearch/fashion-mnist}.
\end{itemize}
The code of the experiments is available at \url{https://github.com/bmnsmets/semitorch}.

\subsection{Architectures}

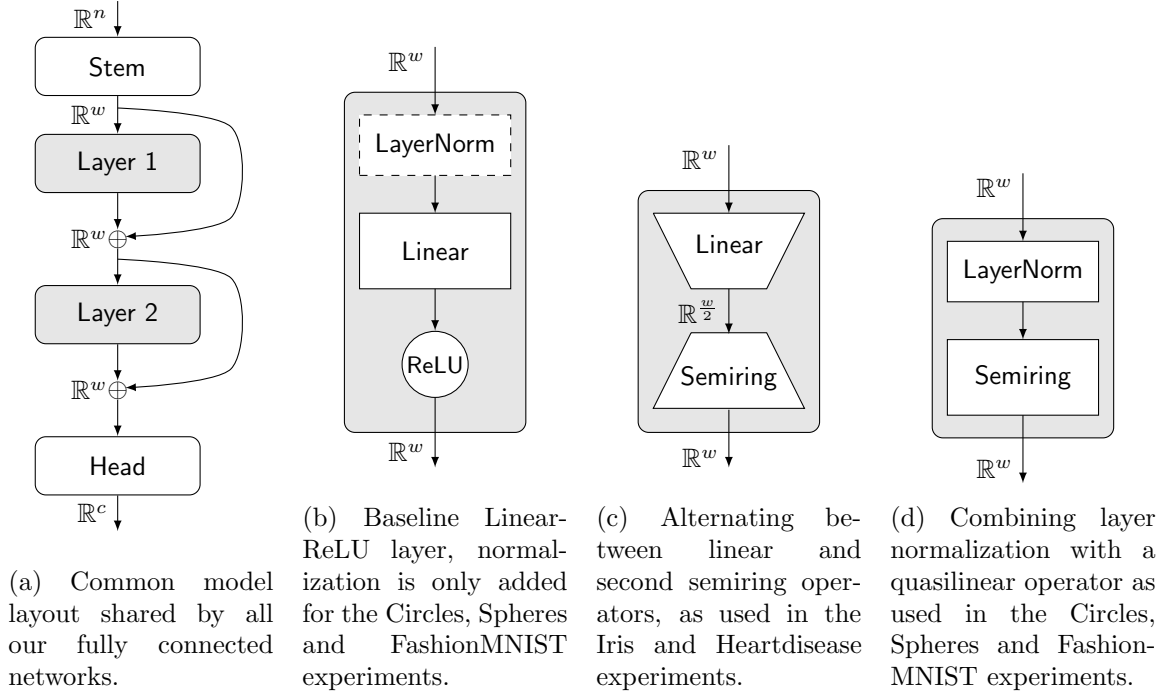
\begin{figure}[ht!]
\centering
    \begin{subfigure}[b]{0.23\textwidth}
        \centering
        \begin{tikzpicture}[every node/.style={font={\small\sffamily}}]
        
        \tikzstyle{block} = [rectangle, draw, text width=5em, text centered, rounded corners, minimum height=2em]
        \tikzstyle{line} = [draw, -latex]
            
        \node [] (input) {};
        \node [block, below of=input] (stem) {Stem};
        \node [block, fill=gray!20, below=0.5cm of stem] (layer1) {Layer 1};
        \node [below of=layer1, inner sep=0pt] (add1) {$\oplus$};
        \node [block, fill=gray!20, below of=add1] (layer2) {Layer 2};
        \node [below of=layer2, inner sep=0pt] (add2) {$\oplus$};
        \node [block, below of=add2] (head) {Head};
        \node [below of=head] (output) {};
        
        \path [line] (input) -- node[left] {$\mathbb{R}^n$} (stem);
        \path [line] (stem) -- node[left] {$\mathbb{R}^w$} (layer1);
        \path [line] (layer1) -- (add1)  node [left] {$\mathbb{R}^w$};
        \path [line] (add1) -- (layer2);
        \path [line] (layer2) -- (add2)  node [left] {$\mathbb{R}^w$};
        \path [line] (add2) -- (head);
        \path [line] (head) --  node [left] {$\mathbb{R}^c$} (output);
        \draw [line] plot [smooth] coordinates {(0,-1.55)  (1.5,-1.8)  (1.5, -3.0)  (0.1,-3.25)};
        \draw [line] plot [smooth] coordinates {(0,-3.55)  (1.5,-3.8)  (1.5, -5.0)  (0.1,-5.25)};
        
        \end{tikzpicture}
        \caption{Common model layout shared by all our fully connected networks.}
        \label{fig:common-architecture}
    \end{subfigure}
    ~
    \begin{subfigure}[b]{0.23\textwidth}
        \centering
        \begin{tikzpicture}[every node/.style={font={\small\sffamily}}]
            \tikzstyle{block} = [rectangle, draw, text width=5em, text centered, rounded corners, minimum height=2em]
            \tikzstyle{line} = [draw, -latex]

            \draw[rounded corners, fill=gray!20] (-1.2,2.1) rectangle (1.2,-2.4);
        	\draw[fill=white, dashed] (-1,1.8) rectangle (1,1.0);
        	\node[] at (0,1.4) {LayerNorm};
        	\draw[fill=white] (-1,0.5) rectangle (1,-0.5);
        	\node[] at (0,0) {Linear};
        	\node[circle, fill=white, draw=black, inner sep=1pt] (relu) at (0,-1.5) {{\footnotesize ReLU}};
        	\node[] (output) at (0,-3.0) {};
        	
        	\path [line] (0,1) -- (0,0.5);
        	\path [line] (0,2.7) -- node[above left] {$\mathbb{R}^w$} (0,1.8);
        	\path [line] (0,-0.5) -- (relu);
        	\path [line] (relu) -- node[below left] {$\mathbb{R}^w$} (output);
        \end{tikzpicture}
        \caption{Baseline Linear-ReLU layer, normalization is only added for the Circles, Spheres and FashionMNIST experiments.}
        \label{fig:relu-layer}
    \end{subfigure}
    ~
    \begin{subfigure}[b]{0.23\textwidth}
        \centering
        \begin{tikzpicture}[every node/.style={font={\small\sffamily}}]
            \tikzstyle{block} = [rectangle, draw, text width=5em, text centered, rounded corners, minimum height=2em]
            \tikzstyle{line} = [draw, -latex]

            \draw[rounded corners, fill=gray!20] (-1.2,0.8) rectangle (1.2,-2.4);
    	\draw[fill=white] (-1,0.5) -- (1,0.5) -- (0.5,-0.5) -- (-0.5,-0.5) -- (-1,0.5);
    	\node[] at (0,0.1) {Linear};
    	\begin{scope}[yshift=-45]
    	\draw[fill=white] (-0.5,0.5) -- (0.5,0.5) -- (1,-0.5) -- (-1,-0.5) -- (-0.5,0.5);
    	\node[] at (0,-0.1) {Semiring};
    	\end{scope}
    	\node[] (output) at (0,-3.0) {};
    	
    	\path [line] (0,1.4) -- node[above left] {$\mathbb{R}^w$} (0,0.5);
    	\path [line] (0,-0.5) -- node[left] {$\mathbb{R}^{\frac{w}{2}}$} (0,-1.1);
    	\path [line] (0,-2.1) -- node[below left] {$\mathbb{R}^w$} (output);
        \end{tikzpicture}
        \caption{Alternating between linear and second semiring operators, as used in the Iris and Heartdisease experiments.}
        \label{fig:semiring-layer-1}
    \end{subfigure}
    ~
    \begin{subfigure}[b]{0.23\textwidth}
        \centering
        \begin{tikzpicture}[every node/.style={font={\small\sffamily}}]
            \tikzstyle{block} = [rectangle, draw, text width=5em, text centered, rounded corners, minimum height=2em]
            \tikzstyle{line} = [draw, -latex]

            \draw[rounded corners, fill=gray!20] (-1.2,2.1) rectangle (1.2,-0.8);
    	\draw[fill=white] (-1,1.8) rectangle (1,1.0);
    	\node[] at (0,1.4) {LayerNorm};
    	\draw[fill=white] (-1,0.5) rectangle (1,-0.5);
    	\node[] at (0,0) {Semiring};
    	
    	\path [line] (0,1) -- (0,0.5);
    	\path [line] (0,2.7) -- node[above left] {$\mathbb{R}^w$} (0,1.8);
    	\path [line] (0,-0.5) -- node[below left] {$\mathbb{R}^w$} (0,-1.4);
        \end{tikzpicture}
        \caption{Combining layer normalization with a quasilinear operator as used in the Circles, Spheres and FashionMNIST experiments.}
        \label{fig:semiring-layer-2}
    \end{subfigure}
\caption{
Network architecture for our fully connected experiments. The {\sffamily Head} and {\sffamily Stem} modules are linear modules. None of the modules include biases. The layer normalization modules include affine transforms.
The number of input features \(n\) and number of output classes \(c\) are dataset dependent, the internal width parameter \(w\) is chosen per experiment.
Each network under consideration has the exact same number of parameters per experiment.
}
\label{fig:fc-architectures}
\end{figure}

We will train a set of network architectures that only slightly vary based on the dataset.
Every model will be based on a common architecture with a linear stem and head with two layers with residual connections in between, see Figure~\ref{fig:common-architecture}.
The stem and head are there to convert the number of features in the dataset \(n\) to a common internal network width \(w\) and back to the number of classes \(c\) in the dataset.
The internal layers we will vary between a traditional ReLU based layer (Figure~\ref{fig:relu-layer}) and semiring based layers (Figure~\ref{fig:semiring-layer-1} and \ref{fig:semiring-layer-2}).
For the iris and heart disease datasets we use the ReLU layer without layer normalization and semiring layer from Figure~\ref{fig:semiring-layer-1}, for the circles, spheres and FashionMNIST dataset we use the ReLU layer with layer normalization and the semiring layer from Figure~\ref{fig:semiring-layer-2}.

\subsection{Training}

Training the networks with semiring based activation proved challenging, the usual setups that are known to work well for conventional neural network do not necessarily carry over.
Indeed, parameter initialization, normalization layers, optimizers, and schedulers are all areas that have seen much research to obtain the best possible results with conventional neural networks.
We spent a lot of time finding a training setup that worked for our modified networks
but we expect that this is an area where more gains could be found.
We document the training setup that we used for our fully-connected experiments next.

\paragraph{Optimizer \& learning rate scheduler.}
We ended up settling on a combination of the \emph{AdamW} optimizer  \citep{loshchilov2019decoupled} and the \emph{1-cycle} learning rate scheduler \citep{smith2019super}.
Key in getting comparable performance out of the semiring based networks as compared to the baseline network was assigning a separate optimizer to the linear parameters and the parameters of the semiring module.
In general we needed to assign smaller learning rates to the semiring parameters to have stable training.
The FashionMNIST experiment is an exception to this, there we obtained best results by increasing the semiring learning rate and decreasing the linear learning rate.
The hyperparameters for all the experiments are listed in Appendix~\ref{app:A}.

\subsection{Initialization}

An important aspect of training neural networks is proper initialization of its parameters.
The standard parameter initialization schemes like \emph{Xavier} initialization \citep{glorot2010understanding} and \emph{Kaiming} initialization \citep{he2015delving} are derived specifically around the forward and backward stability of linear maps and so do not apply to semiring-based maps.
Consequently, we need to come up with new initialization schemes for the semiring weights, specifically for tropical and logarithmic cases.

\paragraph{Tropical.}

For tropical operators (i.e. max-plus and min-plus) we propose an initialization scheme based on `fair' backpropagation of gradients.
Consider the max-plus operator \(R_\text{max}^m \to R_\text{max}^n\) given by
\begin{equation}
    y_i = \max_{j=1\ldots m} w_{ij} + x_j
    \label{eq:tropical-op}
\end{equation}
where \(x_1,\ldots,x_m \in R_\text{max}\) are the inputs, \(y_1,\ldots,y_n \in R_\text{max}\) are the outputs and \([w_{ij}]_{ij} \in R_\text{max}^{n \times m}\) are the trainable parameters.
Then the partial derivatives are given by
\begin{equation*}
    \frac{\partial y_i}{\partial x_j}
    =
    \begin{cases}
        \ 1 \qquad & \text{if } j = \argmax_{k=1 \ldots m} w_{ik} + x_k,
        \\
        \ 0 & \text{else},
    \end{cases}
\end{equation*}
for \(i = 1 \ldots n\) and \(j = 1 \ldots m\).
This could be problematic for the backward pass since if an input \(x_j\) never `wins' one of the maxima, i.e. \(\frac{\partial y_i}{\partial x_j} = 0\) for all \(i=1,\ldots,m\), then its gradient will always be zero.
Conversely, if an input \(x_j\) happens to be very large on a consistent basis then it will accumulate all the gradients of all the outputs \(y_1,\ldots,y_n\) to itself.
The result would be a very unbalanced gradient distribution during the backward pass, something we know from previous research on parameter initialization \citep{glorot2010understanding,he2015delving} is undesirable.

The extreme case for the operator \eqref{eq:tropical-op} would be having an \(x_1\) value (for example) that is consistently much larger than any other \(x_j + w_{ij}\) so that for all \(i = 1,\ldots,n\) we always have \(y_i = w_{i1}+x_1\).

Of course, we have limited control over the degree that this effect will manifest during training, but we can at least avoid it at the start by choosing an appropriate initialization scheme.
The idea is to make sure that at initialization there is a high probability that there is at least one \(i = 1,\ldots,n\) so that \(\frac{\partial y_i}{\partial x_j} = 1\) for each \(j=1,\ldots,m\).
This is of course only possible when \(n \geq m\), but in all our experiments we have used models for which this is the case (see Figure~\ref{fig:fc-architectures} and \ref{fig:convnext}).
Our aim will be to initialize the weights \(w_{ij}\) in \eqref{eq:tropical-op} so that there is a high probability of each of the \(n\) inputs `winning' roughly \(\frac{n}{m}\) of the \(m\) outputs.

Assuming the inputs \(x_j\) generally stay in a range \(\left[-\frac{K}{2},\frac{K}{2}\right]\) for some \(K>0\), then in the max-plus case we can initialize the weight matrix \(W = [w_{ij}]_{ij} \in \mathbb{R}^{m \times n}\) as
\begin{equation}
    w_{ij} = \operatorname{Unif}[-\varepsilon,\varepsilon]
    +
    \begin{cases}
        \ 0 \qquad & \text{if } i = j \bmod m
        \\
        \ -K & \text{else.}
    \end{cases}
    \label{eq:maxplus-init}
\end{equation}
The second term applies a penalty of \(- K\) to each input unless \({i = j \mod m}\) ensuring that the \(m\) available `wins' are fairly distributed amount the \(n\) inputs (at least with high probability based on our assumptions on the inputs).
We additionally add a modest uniform distribution \(\operatorname{Unif}[-\varepsilon,\varepsilon]\) to keep the initialization scheme from being deterministic from run to run, \(\varepsilon\) is chosen on the order of \(\frac{K}{2}\).

A matrix initialized like this, but without the stochastic term, looks like
\begin{equation}
    \begin{bmatrix}
        0 & -K & \cdots & -K
        \\
        -K & 0 & & -K
        \\
        \vdots  & & \ddots &
        \\
        -K & & & 0
        \\
        0 & -K & \cdots & -K
        \\
        -K & 0 & & -K
        \\
        \vdots  & & \ddots &
    \end{bmatrix}
    .
    \label{eq:tropical-fair-matrix}
\end{equation}
We see that in each row we get a single zero coefficient with the others having the value \(-K\).
Assuming that the inputs are in the range \(\left[-\frac{K}{2},\frac{K}{2}\right]\), and we forget the stochastic term for the moment, then the input corresponding to the column with the zero value will achieve the maximum for the output corresponding with that row.
This has the effect of  \((m \bmod n)\) number of inputs contributing to \(\left\lfloor \sfrac{m}{n} \right\rfloor\) outputs and the remaining inputs contributing to \(\left\lceil \sfrac{m}{n} \right\rceil\) outputs.
This avoids the vanishing and exploding gradient problem that would be caused by a single input dominating.

For the min-plus case we similarly set
\begin{equation}
    w_{ij} = \operatorname{Unif}[-\varepsilon,\varepsilon]
    +
    \begin{cases}
        \ 0 \qquad & \text{if } i = j \bmod m
        \\
        \ K & \text{else.}
    \end{cases}
    \label{eq:minplus-init}
\end{equation}
In our experiments normalization layers and weight decay keep input values fairly small, so our starting assumption of inputs being in a range \(\left[-\frac{K}{2},\frac{K}{2}\right]\) generally holds.
We found \(K=1\) gave the intended effect of an initial fair distribution during the backward pass without overly biasing the initialization.
We call this scheme \emph{fair tropical initialization}.

\paragraph{Logarithmic.}

For the logarithmic semiring (for some choice of \(\mu \in \mathbb{R} \setminus \{0\}\)) consider the operator \((R_{\log}^\mu)^m \to (R_{\log}^\mu)^n\) given by
\begin{equation}
    y_i
    =
    \frac{1}{\mu} \log\left(
        \sum_{j=1}^m
        e^{\mu(w_{ij}+x_j)}
    \right)
    \label{eq:log-op}
\end{equation}
for all \(i=1\ldots n\), where \(x_1,\ldots,x_m \in R_{\log}^\mu\) are the inputs, \(y_1,\ldots,y_n \in R_{\log}^\mu\) are the outputs and \([w_{ij}]_{ij} \in (R_{\log}^\mu)^{n \times m}\) are the trainable parameters.
Then the partial derivatives are given by
\begin{equation*}
    \frac{\partial y_i}{\partial x_j}
    =
    \frac{e^{\mu (x_j + w_{ij})}}{\sum_{k=1}^n e^{\mu (x_k + w_{ik})}}
    ,
\end{equation*}
which is nothing but the \emph{softmax} function with temperature \(\frac{1}{\mu}\) over the values \(\{x_k + w_{ik}\}_{k=1}^m\).

Let us now take the same approach as \citet{glorot2010understanding} and look at the inputs \(x_j\) and outputs \(y_i\) as random variables and control forward variance.
We assume all the inputs \(x_j\) are i.i.d. with expected value \(\mathbb{E}[x_j]=0\) and some finite variance \(\Var(x) < \infty\).
Then we can estimate the variance of the outputs using the \emph{delta method} as
\begin{equation*}
    \Var(y_i)
    \approx
    \sum_{j=1}^m \frac{\partial y_i}{\partial x_j}
    \left( \mathbb{E}[x_1],\ldots, \mathbb{E}[x_m] \right)^2 \Var(x_j),
\end{equation*}
and since we assumed the input distributions to be i.i.d. and centered we have
\begin{equation*}
    \Var(y_i)
    \approx
    \Var(x)
    \sum_{j=1}^m \frac{\partial y_i}{\partial x_j}(0,\ldots,0)^2
    =
    \Var(x)
    \sum_{j=1}^m 
    \left( \frac{e^{\mu w_{ij}}}{\sum_{k=1}^m e^{\mu w_{ik}}} \right)^2
    .
\end{equation*}
Ideally we want \(\Var(y_i) \approx \Var(x)\), which is the case if
\begin{equation*}
    \sum_{j=1}^m
    \left( \frac{e^{\mu w_{ij}}}{\sum_{k=1}^m e^{\mu w_{ik}}} \right)^2
    \approx
    1
\end{equation*}
or
\begin{equation}
    \sum_{j=1}^m \left( e^{\mu w_{ij}} \right)^2
    \approx
    \left( \sum_{j=1}^m e^{\mu w_{ij}} \right)^2
    \label{eq:log-forward-condition}
\end{equation}
for all \(i=1,\ldots,n\).
We can satisfy this condition exactly by choosing a single \(J_i  = 1,\ldots,m\) for each \(i=1,\ldots,n\) and set that weight to \(w_{i,J_i} = 0\) and set the other weights to \(w_{ij} = - \sgn(\mu) \infty\).
In that case \(e^{\mu w_{ij}} = 1\) if \(j = J_i\) and \(e^{\mu w_{ij}} = 0\) if not and consequently both sides of \eqref{eq:log-forward-condition} are equal to \(1\).

We can also make a similar analysis for the backward pass.
Let \((\overline{y}_i)_{i=1}^n\) be the loss gradients of the outputs and \((\overline{x}_j)_{j=1}^m\) be the loss gradients of the inputs.
We interpret these as random variables, where we assume the output gradients \(\overline{y}_i\) are i.i.d. with \(\mathbb{E}[\overline{y}_i]=0\) and finite variance \(\Var(\overline{y}_i)=\Var(\overline{y})<\infty\) for all \(i=1,\ldots,n\).
Then the loss gradients of the inputs are computed as
\begin{equation*}
    \overline{x}_j
    =
    \sum_{i=1}^n \frac{\partial y_i}{\partial x_j}(x_1,\ldots,x_m)
    \overline{y}_i
    ,
\end{equation*}
or if we are talking about the expected backward pass over the possible inputs we can say
\begin{equation*}
    \overline{x}_j
    \approx
    \sum_{i=1}^n \frac{\partial y_i}{\partial x_j}(0,\ldots,0)
    \overline{y}_i
    .
\end{equation*}
The variance of the input gradients can then be approximated as
\begin{equation*}
    \Var(\overline{x}_j)
    \approx
    \Var\left(
    \sum_{i=1}^n \frac{\partial y_i}{\partial x_j}(0,\ldots,0)
    \overline{y}_i
    \right)
    =
    \Var(\overline{y}) 
    \sum_{i=1}^n \left( \frac{e^{\mu w_{ij}}}{\sum_{k=1}^n e^{\mu w_{ik}}} \right)^2
    ,
\end{equation*}
which implies that to get \(\Var(\overline{x}_j) \approx \Var(\overline{y})\) we need
\begin{equation*}
    \sum_{i=1}^n \left( \frac{e^{\mu w_{ij}}}{\sum_{k=1}^n e^{\mu w_{ik}}} \right)^2
    \approx
    1.
\end{equation*}
or
\begin{equation}
    \sum_{i=1}^n \left( e^{\mu w_{ij}} \right)^2
    \approx
    \left( \sum_{i=1}^n e^{\mu w_{ij}} \right)^2
    \label{eq:log-backward-condition}
\end{equation}
for all \(j=1,\ldots,m\).
We can satisfy this condition exactly by choosing a single \(I_j  = 1,\ldots,n\) for each \(j=1,\ldots,m\) and set that weight to \(w_{I_j,j} = 0\) and set the other weights to \(w_{ij} = - \sgn(\mu) \infty\).
In that case \(e^{\mu w_{ij}} = 1\) if \(i = I_j\) and \(e^{\mu w_{ij}} = 0\) if not and consequently both sides of \eqref{eq:log-backward-condition} are equal to \(1\).

Satisfying both forward \eqref{eq:log-forward-condition} and backward condition \eqref{eq:log-backward-condition} is only possible in the case that \(m=n\) where we can set \(I_j = j\) and \(J_i = i\).
But even then, this scheme is not satisfactory.
If we initialize parameters to \(\pm \infty\) (depending on the sign of \(\mu\)) they can not be changed by the addition of finite numbers thus making training impossible.
We might then be tempted to substitute \(\pm\infty\) by a very large negative of positive value to initialize with.
But this still causes a problem for training.

Consider the loss gradient for an individual parameter:
\begin{equation}
    \overline{w}_{ij}
    =
    \frac{e^{\mu (x_j + w_{ij})}}{\sum_{k=1}^m e^{\mu (x_k + w_{i k})}}
    \overline{y}_i
    .
    \label{eq:log-parameter-gradient}
\end{equation}
If we try to satisfy \eqref{eq:log-forward-condition} and \eqref{eq:log-backward-condition} then we would set \(w_{i k} = 0\) for some \(k = 1,...,n\) for every \(i=1,\ldots,n\).
This implies that the denominator in \eqref{eq:log-parameter-gradient} is greater than or equal to \(1\).
If we subsequently set \(w_{i j} = - \sgn{(\mu)} K \) for \(j \neq k\) and \(K\) a very large number then the numerator of \eqref{eq:log-parameter-gradient} becomes vanishingly small, possible zero in floating point format.
This essentially freezes the parameter's value since any updates applied to it would become practically zero.

Assuming the inputs \((x_j)_{j=1}^m\) are centered, the expected value of the fraction in \eqref{eq:log-parameter-gradient} is given by
\begin{equation}
    \frac{e^{\mu w_{ij} }}{\sum_{k=1}^m e^{\mu w_{i k} }},
    \label{eq:log-parameter-factor}
\end{equation}
which gives values in the range \([0,1]\) for any \(i=1,\ldots,n\) and \(j=1,\ldots,m\).
We can avoid this fraction becoming (effectively) zero from the start by initializing all the parameters to roughly the same value, i.e. for all \(i=1,\ldots,n\) set 
\begin{equation}
    w_{i 1} \approx w_{i 2} \approx \ldots \approx w_{i m}.
    \label{eq:log-parameter-condition}
\end{equation}
In this case we achieve the \emph{maximum of the minimum} where \eqref{eq:log-parameter-factor} gives the value \(\sfrac{1}{m}\) for all \(i=1,\ldots,n\) and \(j=1,\ldots,m\).

Satisfying the forward condition \eqref{eq:log-forward-condition}, backward condition \eqref{eq:log-backward-condition} and parameter trainability condition \eqref{eq:log-parameter-condition} all at the same time is not possible.
However, the fair tropical initialization scheme does present a reasonable trade-off between these three clashing requirements.
Indeed if we initialize using this scheme (assuming \(n>m\) for the moment) we get a matrix like \eqref{eq:tropical-fair-matrix},
where we let \(K>0\) if \(\mu>0\) and \(K<0\) if \(\mu<0\).
First, most entries have the same \(-K\) value and so satisfy \eqref{eq:log-parameter-condition} to some degree.
Second, in every row we have a single element (the zero values) that are relatively dominant with respect to the other elements in the row, thus catering to \eqref{eq:log-forward-condition}.
Third, in every column we have at most \(\lceil \sfrac{n}{m} \rceil\) zero values that are relatively dominant with respect to the other elements in the column, thus catering to \eqref{eq:log-backward-condition}. 

\subsection{Results}

We perform 10 training runs for each type of network on each dataset using the training setup we described.
We measure performance by the accuracy (mean \(\pm\) standard deviation over the runs) on the testing dataset that was not seen during training.
We use the same training/testing split of the dataset for every type of network.
The results are summarized in Table~\ref{tbl:fullyconnected}.

\begin{table}[ht]
\begin{tabular}{lllllll}
Model / dataset & Iris & Heartdisease & Circles & Spheres & FashionMNIST
\\
\hline
ReLU    
& $97.14\ {\scriptstyle\pm0.62}$ 
& ${\color{Plum}\bm{83.93}}\ {\scriptstyle\pm2.16}$ 
& $84.50\ {\scriptstyle\pm0.39}$
& $80.91\ {\scriptstyle\pm1.62}$
& ${\color{Plum}\bm{83.82}}\ {\scriptstyle\pm0.35}$
\\[0.5em]
maxplus 
& $97.52\ {\scriptstyle\pm1.01}$ 
& ${\color{Green}83.50}\ {\scriptstyle\pm2.27}$ 
& $84.84\ {\scriptstyle\pm0.86}$
& ${\color{Plum}\bm{81.69}}\ {\scriptstyle\pm0.42}$
& $83.50\ {\scriptstyle\pm0.34}$
\\
minplus 
& $97.62\ {\scriptstyle\pm0.49}$ 
& $82.84\ {\scriptstyle\pm1.22}$ 
& $84.91\ {\scriptstyle\pm0.35}$
& $81.61\ {\scriptstyle\pm0.60}$
& $83.39\ {\scriptstyle\pm0.24}$
\\[0.5em]
logplus $\mu=-10$ 
& $97.58\ {\scriptstyle\pm1.00}$ 
& $81.72\ {\scriptstyle\pm1.62}$ 
& ${\color{Green}85.06}\ {\scriptstyle\pm0.25}$
& $81.52\ {\scriptstyle\pm0.71}$
& $83.46\ {\scriptstyle\pm0.37}$
\\
logplus $\mu=-1$ 
& ${\color{Green}97.90}\ {\scriptstyle\pm0.39}$ 
& $83.26\ {\scriptstyle\pm2.15}$ 
& $73.92\ {\scriptstyle\pm6.84}$
& $69.41\ {\scriptstyle\pm5.05}$
& $83.50\ {\scriptstyle\pm0.28}$
\\
logplus $\mu=1$ 
& ${\color{Plum}\bm{97.97}}\ {\scriptstyle\pm0.49}$ 
& $82.38\ {\scriptstyle\pm1.73}$ 
& $75.06\ {\scriptstyle\pm7.89}$
& $67.28\ {\scriptstyle\pm5.53}$
& $83.46\ {\scriptstyle\pm0.16}$
\\
logplus $\mu=10$ 
& $97.46\ {\scriptstyle\pm0.62}$ 
& $81.86\ {\scriptstyle\pm1.15}$ 
& ${\color{Plum}\bm{85.16}}\ {\scriptstyle\pm0.26}$
& ${\color{Green}81.62}\ {\scriptstyle\pm0.45}$
& ${\color{Green}83.56}\ {\scriptstyle\pm0.37}$
\\
\hline
Parameters & 60 & 5328 & 2336 & 2336 & 2288
\end{tabular}
\caption{
Accuracy (mean $\pm$ standard deviation) of the trained fully connected networks on the testing sets of the various classification datasets.
The best result for each dataset is indicated in {\color{Plum}\textbf{purple}}, the second best result in {\color{Green}green}.
}
\label{tbl:fullyconnected}
\end{table}

The semiring activation networks manage to modestly outperform the classic network in \(3\) of the \(5\) cases and are only slightly behind in the other \(2\) cases.
A standout failure is the logarithmic semiring networks with \(\mu \in \{-1,1\}\) in the \emph{circles} and \emph{spheres} datasets.
We can explain this based on the fact that these are low-dimensional problem (2 respectively 3 input dimensions) where there is a sharp boundary between the classes.
At the same time the logarithmic maps are fairly gradual for small absolute values of \(\mu\) and weight decay keeps the parameters from becoming large enough to compensate for that.
Consequently, the logarithmic networks have a hard time separating classes that are close together.

On the whole though we can conclude that the semiring approach is at least as viable as the standard approach for these smaller problems.
We can now try to see if we can scale up to some larger networks.

\section{In convolutional neural networks}

We perform a final experiment to see how viable replacing activation functions with semiring activation is in an existing modern and optimized network architecture.
We start from the \emph{ConvNeXt} network from \citet{liu2022convnet}, specifically the \emph{atto} variant available from \citet{wightman2019timm}, and train it on FashionMNIST.

\subsection{Architecture modifications}

\begin{figure}[ht!]
\centering
    \begin{subfigure}[t]{0.4\textwidth}
        \centering
        \resizebox{0.6\textwidth}{!}{
        \begin{tikzpicture}
        \tikzstyle{block} = [rectangle, draw, text width=5em, text centered, minimum height=2em]
        \tikzstyle{trapez} = [draw, text width=2em, text centered, minimum height=2.5em]
        \tikzstyle{line} = [draw, -latex]
        
        \node [] (input) {};
        \node [block, below=0.75cm of input] (dconv) {Depthwise Conv};
        \node [trapez, trapezium, trapezium angle=60, fill=gray!20, below=0.75cm of dconv] (layer1) {Linear};
        \node [rectangle, draw, fill=gray!20, below=0.5cm of layer1, inner sep=4pt] (act) {\small GELU};
        \node [trapez, trapezium, trapezium angle=-60, below=0.5cm of act] (layer2) {Linear};
        \node [circle,below=0.5cm of layer2, inner sep=0pt] (add2) {$\oplus$};

        \path [line] (input) -- node[left] {$\mathbb{R}^{h \times w \times c}$} (dconv);
        \path [line] (dconv) -- node[left] {$\mathbb{R}^{h \times w \times c}$} (layer1);
        \path [line] (layer1) -- node[left]{$\mathbb{R}^{h \times w \times 4 c}$} (act);
        \path [line] (act) -- node[left]{$\mathbb{R}^{h \times w \times 4 c}$} (layer2);
        \path [line] (layer2) -- (add2) ;
        \path [line] (add2) -- node [left] {$\mathbb{R}^{h \times w \times c}$} ++ (0,-0.75)  ;
        
        \draw [line] plot [smooth, tension=0.2] coordinates {(0,-0.4)  (1.8,-0.8)  (1.8,-6.5)  (0.2,-6.85)};
        \end{tikzpicture}
        }
        \caption{Standard ConvNeXt block with two linear maps and an activation function in between.}
        \label{fig:std-convnext-block}
    \end{subfigure}
    \hspace{4em}
    \begin{subfigure}[t]{0.4\textwidth}
        \centering
        \resizebox{0.6\textwidth}{!}{
        \begin{tikzpicture}
        \tikzstyle{block} = [rectangle, draw, text width=5em, text centered, minimum height=2em]
        \tikzstyle{trapez} = [draw, minimum width=2em, text centered, minimum height=2.5em, inner sep=8pt]
        \tikzstyle{line} = [draw, -latex]
            
        \node [] (input) {};
        \node [block, below=0.75cm of input] (dconv) {Depthwise Conv};
        \node [trapez, trapezium, trapezium angle=60, fill=gray!20, below=0.75cm of dconv, align=center] (layer1) {Semiring};
        \node [trapez, trapezium, trapezium angle=-60, below=1.46cm of layer1] (layer2) {Linear};
        \node [circle,below=0.5cm of layer2, inner sep=0pt] (add2) {$\oplus$};

        \path [line] (input) -- node[left] {$\mathbb{R}^{h \times w \times c}$} (dconv);
        \path [line] (dconv) -- node[left] {$\mathbb{R}^{h \times w \times c}$} (layer1);
        \path [line] (layer1) -- node[left]{$\mathbb{R}^{h \times w \times 4 c}$} (layer2);
        \path [line] (layer2) -- (add2) ;
        \path [line] (add2) -- node [left] {$\mathbb{R}^{h \times w \times c}$} ++ (0,-0.75)  ;
        
        \draw [line] plot [smooth, tension=0.2] coordinates {(0,-0.4)  (1.8,-0.8)  (1.8,-6.35)  (0.2,-6.75)};
        \end{tikzpicture}
        }
        \caption{Modified ConvNeXt block with a semiring and linear layer without activation function.}
        \label{fig:modified-convnext-block}
    \end{subfigure}
\caption{
Standard and semiring-based ConvNeXt \citep{liu2022convnet} blocks compared. 
Normalization and dropout modules are omitted.
}
\label{fig:convnext}
\end{figure}
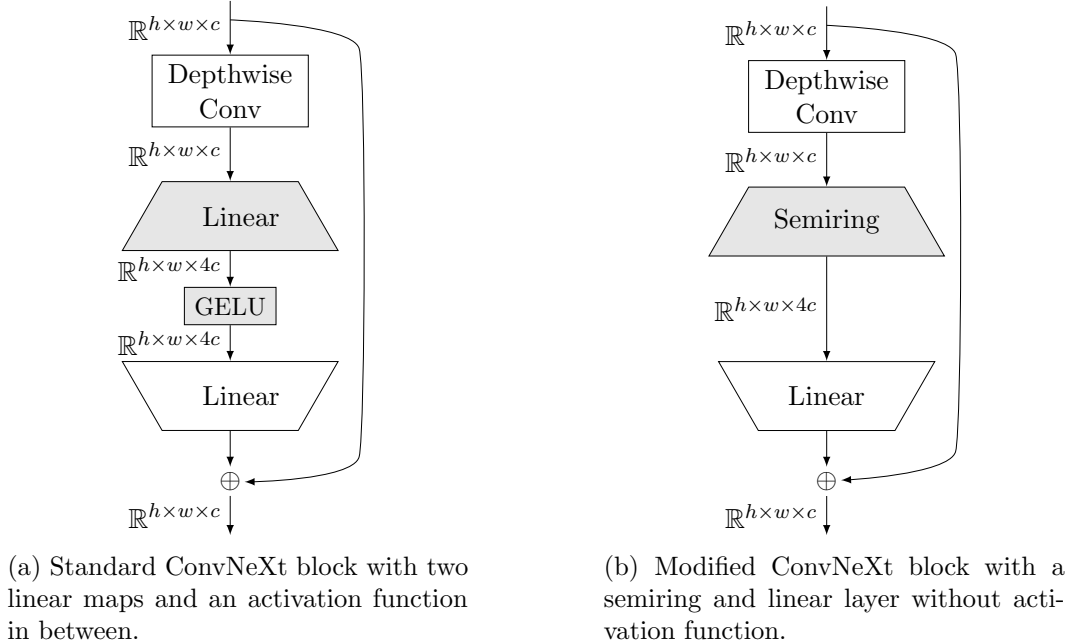

The core \emph{block} at the heart of a ConvNeXt consists of two phases: 
\begin{enumerate}[noitemsep]
    \item a \emph{depthwise convolution} that applies a kernel per feature map and
    \item a \emph{normalization} followed by a \emph{reverse bottleneck MLP} that is applied \emph{pixel-wise},
\end{enumerate}
the result of these two operations is added to the input to form a residual connection, see Figure~\ref{fig:std-convnext-block}.

Our adaptation keeps the (linear) depthwise convolution and normalization but replaces the MLP with a concatenation of a non-linear semiring operator and a linear operator.
The semiring operation does a \(4 \times\) fan-out while the linear operation reduces again to the original amount of channels, thus retaining the same reverse bottleneck of the original, see Figure~\ref{fig:modified-convnext-block}.

\subsection{Results}

We compare the performance of the semiring based networks against the baseline network using an MLP.
We perform 10 training runs per model and record accuracy on the test dataset.
We train 4 semiring based models: maxplus, minplus, logplus(\(\mu=-1\)) and logplus(\(\mu=1\)).
We train all models for the same 50 epochs and batchsize of 512 and adapt the rest of the training setup to each model.
Details for the training setup can be found in Appendix~\ref{app:A}.
The results are summarized in Table~\ref{tbl:convnext-results}.

Training with these (relatively) large scale networks proved more challenging than with the previous toy networks.
In particular the logarithmic networks proved challenging and required us to remove the affine transform from the normalization layer to avoid numeric stability issues.
The problem being that in single precision floating point, the function \(x \mapsto e^x\) already overflows at \(x = 89\).
This would not happen if the input \(x\) is normalized but is likely enough to happen if we let \(x\) be an affine transform of a normalized input.
The overflow would only need to happen at one place in the network, after which `inf' values would propagate throughout the network and ruin training.
As we already experienced in the previous experiments, both the tropical and logarithmic variants proved to be more sensitive to the training hyperparameters than the baseline network.
This sensitivity makes finding a good training setup harder for the semiring networks than for the baseline network.

\begin{table}[ht!]
    \centering
    \begin{tabular}{llll}
    Feed-forward type & Test accuracy(\%) & Train accuracy(\%) & Gap(\%)
    \\
    \hline
    MLP (linear-GELU-linear) & \(91.24\ {\scriptstyle\pm0.13}\) & \(99.97\ {\scriptstyle\pm0.07}\) & \(-8.63\)
    \\
    maxplus-linear & \(89.08\ {\scriptstyle\pm0.29}\) & \(93.26\ {\scriptstyle\pm2.32}\) & \(-4.18\)
    \\
    minplus-linear & \(89.93\ {\scriptstyle\pm0.24}\) & \(94.74\ {\scriptstyle\pm2.79}\) & \(-4.81\)
    \\
    logplus-linear \(\mu=-1\) \  & \(88.15\ {\scriptstyle\pm0.15}\) & \(91.72\ {\scriptstyle\pm2.52}\) & \(-3.57\)
    \\
    logplus-linear \(\mu=1\) \  & \(88.31\ {\scriptstyle\pm0.23}\) & \(91.60\ {\scriptstyle\pm2.52}\) & \(-3.29\)
    \end{tabular}
    \caption{Accuracy(\%) (mean \(\pm\) standard deviation) of the trained ConvNeXt models on the FashionMNIST test dataset and on the last 100 training batches. The generalization gap is the difference between the two mean accuracies.}
    \label{tbl:convnext-results}
\end{table}

As the second column of Table~\ref{tbl:convnext-results} shows, both the tropical and logarithmic network's accuracy falls significantly short of the baseline network.
This shortfall does not bode well for the semiring idea but there are some nuances to be made.

Looking at the third column of Table~\ref{tbl:convnext-results} we see that the baseline network has saturated its performance on the training data.
During the last 100 batches the baseline network has virtually \(100\%\) training accuracy, consequently it can not benefit much more from further training.
At the same time there is a fairly large gap between the training and testing accuracy as can be seen in the last column of Table~\ref{tbl:convnext-results}.
On the side of the semiring networks we see that after 50 epochs there is still significant room for improvement on the training data and that the gap between the testing and training accuracy is much more modest.

We conjecture that this difference is partly explained by the standard training regime being very well suited for the baseline network but that the modifications we made---chiefly the parameter initialization scheme---are not sufficient to extract maximum performance from the semiring networks.
Indeed, the methods for neural network training have evolved much over the last decade with much research into optimizers, schedulers, initialization, regularization, etc., all focused on the linear with activation type networks.
It would not be unreasonable to assume similar efforts could yield similar progress in training semiring based networks.

\section{Discussion \& Concluding remarks}

\paragraph{Viability.}
In this article we have proposed a general framework for constructing trainable nonlinearities for neural networks.
We constructed several such trainable nonlinearities based on the tropical and logarithmic semirings and introduced an associated parameter initialization scheme.
We did a series of experiments to show the viability of replacing the traditional activation function in neural networks with nonlinearities based on the aforementioned semirings.

\paragraph{Unrealized potential.}
Our experiments showed that the semiring approach is viable in that we can get very good results for small networks and decent but not state-of-art results for larger networks.
With regard to the larger scale experiment we concluded that there is more performance on the table for semiring networks that can potentially be unlocked by designing more suitable optimizers, schedulers, initialization schemes, regularization schemes, etc.
Whether such a line of research would be worthwhile is debatable for two reasons.

First, while our experiments show viability, they do not immediately show a clear advantage over traditional neural networks with activation functions.

Second, the current development of deep learning hardware \citep{dhilleswararao2022efficient} focuses to a large degree on optimizing linear operations. 
Most GPUs in general use today already contain hardware dedicated to linear matrix multiplication \citep{markidis2018nvidia}, this makes linear computations much more efficient than doing semiring computations that would have to be executed by general purpose computing units at a higher cost in time and energy.

\paragraph{Conclusion on PDE-G-CNNs.}
Recall that our interest in the semiring structure was instigated by our previous research into PDE-G-CNNs \citep{smets2023pde}.
In those networks we also had trainable nonlinear operators based on the tropical semiring, but these operators were further structured to satisfy equivariant and PDE properties.
We asked whether the benefits from PDE-G-CNNs were perhaps not largely a consequence of the tropical semiring structure rather than the equivariant and PDE structures.

We can now answer this question in the negative.
Without the equivariance and PDE constraints the tropical operator becomes very hard to deal with and a challenge to train.
PDE-G-CNNs can be successfully trained without much adaptation of the training regime and are not unusually sensitive to hyperparameters.
PDE-G-CNNs also do not require more epochs to saturate their training data than normal CNNs, which was clearly an issue for the tropical semiring networks in our ConvNeXt experiment.

Overall, we conclude that the equivariant and PDE structures play an important role in making PDE-G-CNNs work and the semiring structure is not---by itself---the cause of the benefits of PDE-G-CNNs.



\newpage

\appendix
\section*{Appendix A. Hyperparameters for the fully connected experiments}
\label{app:A}

\begin{table}[ht]
\centering
\resizebox{1\textwidth}{!}{
\begin{tabular}{lrrrrrr}
Hyperparameter & Iris & Heartdisease & Circles & Spheres & FashionMNIST
\\
\hline
Epochs & 40 & 40 & 100 & 100 & 40
\\
Batchsize & 8 & 16 & 32 & 16 & 512
\\
Optimizer & AdamW & AdamW & AdamW & AdamW & AdamW
\\
Scheduler & 1-cycle cos & 1-cycle cos & 1-cycle cos & 1-cycle cos & 1-cycle cos
\\
Learning rate (linear) & $0.020$ & $0.010$ & $0.020$ & $0.020$ & $0.008$
\\
Learning rate (tropical) & $0.004$ & $0.008$ & $0.010$ & $0.010$ & $0.040$
\\
Learning rate (logarithmic) & $0.040$ & $0.008$ & $0.008$ & $0.008$ & $0.040$
\\
Weigh decay & $0.01$ & $0.05$ & $0.01$ & $0.01$ & $0.01$
\\
Warmup epochs & 18 & 18 & 45 & 45 & 18
\\
Warmup factor & \(\sfrac{1}{10}\) & \(\sfrac{1}{10}\) & \(\sfrac{1}{10}\) & \(\sfrac{1}{10}\) & \(\sfrac{1}{10}\)
\\
Annihilation factor & \(\sfrac{1}{1000}\) & \(\sfrac{1}{1000}\) & \(\sfrac{1}{1000}\) & \(\sfrac{1}{1000}\) & \(\sfrac{1}{1000}\)
\\
Internal width \(w\) & $4$ & $48$ & $16$ &  $32$ & $8$
\\
RNG seed & 42 & 42 & 42 & 42 & 42
\\
\hline
Parameters & 60 & 5328 & 2336 & 2336 & 2288
\end{tabular}
}
\caption{
Dataset dependent hyperparameters for the fully connected experiments.
}
\label{tbl:fc-learningrates}
\end{table}

\section*{Appendix B. Hyperparameters for the ConvNeXt experiments}
\label{app:B}
\begin{table}[ht!]
    \centering
    \begin{tabular}{lrrrrr}
         Hyperparameter & MLP & tropical-linear & logplus-linear 
         \\
         \hline
         Epochs & 50 & 50 & 50
         \\
         Batchsize & 512 & 512 & 512
         \\
         Affine LayerNorm & Yes & Yes & No
         \\
         Optimizer & AdamW & AdamW & AdamW
         \\
         Scheduler & 1-cycle cos & 1-cycle cos & 1-cycle cos
         \\
         Learning rate (linear) & \(4 \cdot 10^{-3}\) & \(4 \cdot 10^{-3}\) & \(6 \cdot 10^{-3}\)
         \\
         Learning rate (semiring) & -  & \(1 \cdot 10^{-3}\) & \(5 \cdot 10^{-4}\)
         \\
         Weight decay (linear) & \(5 \cdot 10^{-3}\) & \(5 \cdot 10^{-3}\) & \(5 \cdot 10^{-3}\)
         \\
         Weight decay (semiring) & - & \(1 \cdot 10^{-2}\) & \(1 \cdot 10^{-2}\)
         \\
         Warmup epochs & 5 & 5 & 5
         \\
         Warmup factor (linear) & \(\sfrac{1}{25}\) & \(\sfrac{1}{50}\) & \(\sfrac{1}{50}\)
         \\
         Warmup factor (semiring) & - & \(\sfrac{1}{50}\) & \(\sfrac{1}{50}\)
         \\
         Annihilation factor & \(\sfrac{1}{500}\) & \(\sfrac{1}{250}\) & \(\sfrac{1}{250}\)
         \\
         RNG seed & 42 & 42 & 42
         \\
         Initialization (linear) & Kaiming & Kaiming & Kaiming
         \\
         Initialization (semiring) & - & Tropical fair & Tropical fair
         \\
         \hline
         Parameters & \(3,375,850\)  & \(3,375,850\) & \(3,372,170\)
    \end{tabular}
    \caption{Hyper parameters settings for the ConvNeXt experiments.}
    \label{tbl:convnext-hyperparameters}
\end{table}


\vskip 0.2in
\bibliography{main}

\end{document}